\documentclass[10pt,twocolumn,letterpaper]{article}

\usepackage[pagenumbers]{cvpr} 
\usepackage{multirow}
\usepackage{multirow}
\usepackage{algorithm}  
\usepackage{algorithmicx}  
\usepackage{algpseudocode}  
\usepackage{array}  
\usepackage[utf8]{inputenc} 
\usepackage[T1]{fontenc}    
\usepackage{url}            
\usepackage{booktabs}       
\usepackage{amsfonts}       
\usepackage{nicefrac}       
\usepackage{microtype}      
\usepackage{lipsum}  
\usepackage{amsmath}
\usepackage{multirow}       
\usepackage{amsthm}
\usepackage{graphicx}

\usepackage{multirow}
\usepackage[table,xcdraw]{xcolor}
\usepackage{graphicx} 
\usepackage{adjustbox} 
\usepackage{newfloat}
\usepackage{listings}

\usepackage{wrapfig}
\usepackage{caption}

\usepackage{float}
\definecolor{cvprblue}{rgb}{0.21,0.49,0.74}
\usepackage[pagebackref,breaklinks,colorlinks,allcolors=cvprblue]{hyperref}
\usepackage{color}

\def\paperID{41168} 
\def\confName{CVPR}
\def\confYear{2026}

\title{MoAPT: Mixture of Adversarial Prompt Tuning for Vision-Language Models}
\author{%
  Shiji Zhao$^{1}$\footnotemark[1], Qihui Zhu$^{1}$\footnotemark[1],  Shukun Xiong$^{1}$,  Shouwei Ruan$^{1}$,  Maoxun Yuan$^{1}$, \\ Jialing Tao, Jiexi Liu, Ranjie Duan, Jie Zhang$^{2}$, Jie Zhang$^{3}$,  Xingxing Wei$^{1}$\footnotemark[2] \\
  $^{1}$Institute of Artificial Intelligence, Beihang University, Beijing, China \\ 
    $^{2}$Center for Frontier AI Research, A*STAR,  Singapore \\  
    $^{3}$	Institute of Computing Technology, Chinese Academy of Sciences, China \\  
  \texttt{\{zhaoshiji123,xxwei\}@buaa.edu.cn} \\
}

\begin{document}
\maketitle
  \renewcommand{\thefootnote}{\fnsymbol{footnote}} 
  \footnotetext[1]{Equal Contribution.} 
\footnotetext[2]{Corresponding Author.} 
\begin{abstract}
Large pre-trained Vision Language Models (VLMs) demonstrate excellent generalization capabilities but remain highly susceptible to adversarial examples, posing potential security risks. To improve the robustness of VLMs against adversarial examples, adversarial prompt tuning methods are proposed to align the text feature with the adversarial image feature without changing model parameters. However, when facing various adversarial attacks, a single learnable text prompt has insufficient generalization to align well with all adversarial image features, which ultimately results in overfitting. 
To address the above challenge, in this paper, we empirically find that increasing the number of learned prompts yields greater robustness improvements than simply extending the length of a single prompt. Building on this observation, we propose an adversarial tuning method named \textbf{Mixture of Adversarial Prompt Tuning (MoAPT)} to enhance the generalization against various adversarial attacks for VLMs. MoAPT aims to learn mixture text prompts to obtain more robust text features. To further enhance the adaptability, we propose a conditional weight router based on the adversarial images to predict the mixture weights of multiple learned prompts, which helps obtain sample-specific mixture text features aligning with different adversarial image features. Extensive experiments across 11 datasets under different settings show that our method can achieve better adversarial robustness than state-of-the-art approaches.
\end{abstract}    

\section{Introduction}
\label{sec:intro}

Large pre-trained Vision Language Models (VLMs) such as CLIP \cite{radford2021learning}  have excellent generalization capabilities and can be regarded as foundation models \cite{bommasani2021opportunities} in different downstream tasks, e.g., image-text retrieval, zero-shot image classification, or image generation guidance. Due to its wide range of application scenarios, it places high requirements on security performance. However, despite its excellent performance, VLMs face many potential security risks \cite{inkawhich2023adversarial,mao2022understanding,schlarmann2023adversarial}, including the fact that visual models are vulnerable to adversarial examples \cite{szegedy2013intriguing}, which can pose a serious threat to the application in actual scenarios.

To eliminate this potential security risk, many works have been proposed to improve the robustness of VLMs to adversarial examples, which can be mainly divided into two types, full-parameter fine-tuning \cite{mao2022understanding,wang2024pre,schlarmann2024robust,yu2024text} and parameter-efficient fine-tuning \cite{mao2022understanding,wang2024pre,zhang2024adversarial,li2024one,zhou2024few,luo2024adversarial}. Among them, full-parameter fine-tuning is an effective method to improve the adversarial robustness of the model. However, this method often requires a lot of computational overhead and also affects the performance of the model on general tasks. Another type of parameter-efficient method, e.g., adversarial prompt tuning \cite{zhang2024adversarial}, freezes all or most of the weights of the model and only fine-tunes some of its parameters. This type of method can also improve the adversarial robustness with lower training overhead compared with full-parameter fine-tuning, which is a promising solution. However, adversarial prompt tuning faces a serious problem: \textbf{insufficient generalization}. For example, for the text prompt tuning \cite{zhang2024adversarial,li2024one}, when only one learnable prompt is fine-tuned, the text feature is not sufficient to fit the image features for various adversarial examples, which can easily lead to overfitting and further cause potential security risks \cite{wang2025tapt}.

\begin{figure*}[t]
  \centering
\includegraphics[width=\textwidth]{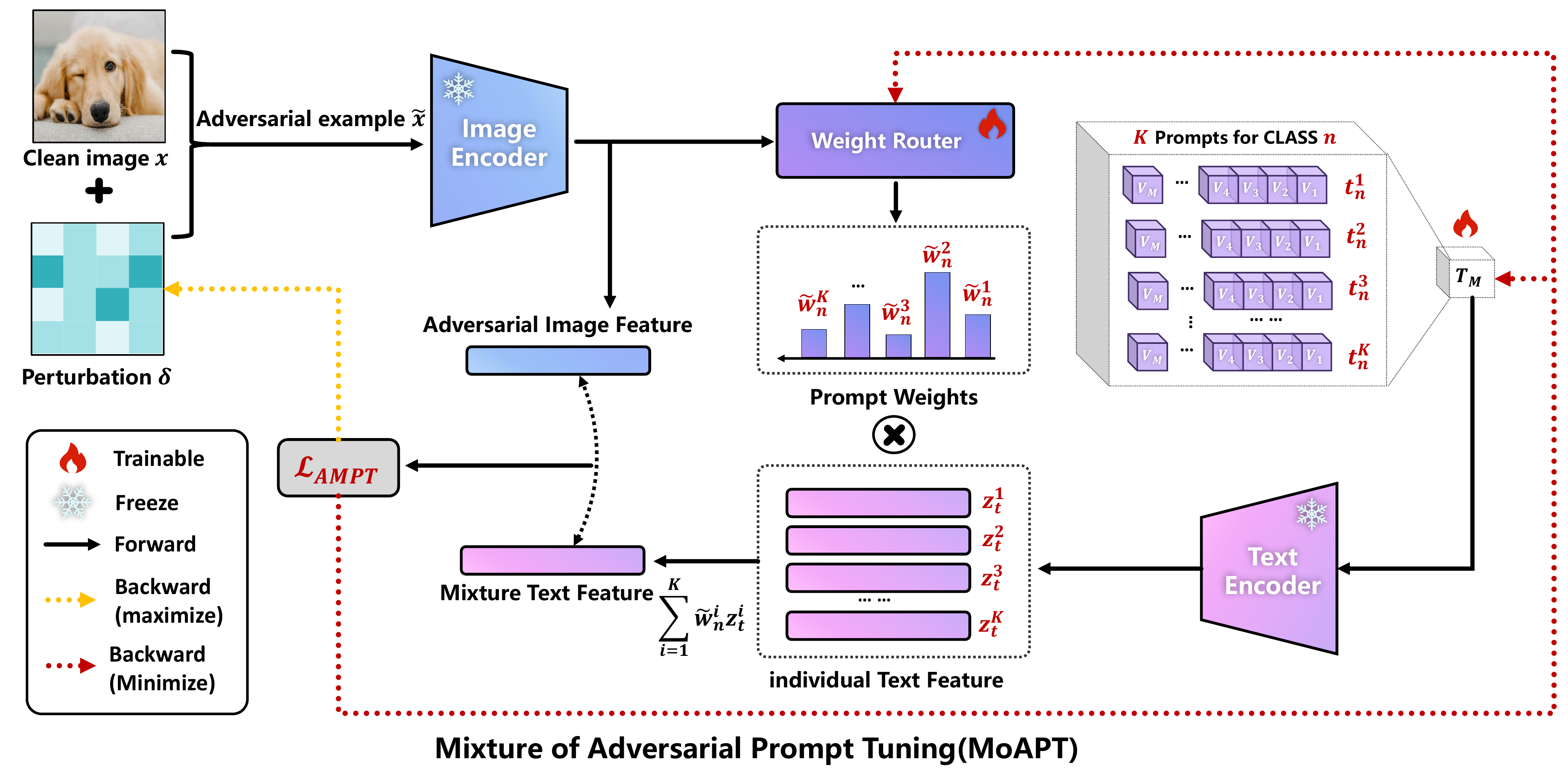}\\
\caption{The framework of \textbf{Mixture of Adversarial Prompt Tuning (MoAPT)}. To enhance the adversarial robustness, we apply adversarial mixture prompt to generate diverse individual text feature, and utilize the conditional prompt weight router to obtain a sample-specific mixture text feature, and finally bring more generalization towards different adversarial examples.}
\label{framework}
\end{figure*}


To enhance the generalization of the adversarial text prompt, an intuitive approach is to increase the length of the text prompt. However, we find that when it grows to a certain length, a longer prompt will bring greater optimization difficulty, and also needs higher requirements on the corresponding text encoder to deal with long prompts, finally leading to suboptimal robustness. Inspired by the Mixture of Experts (MoE) paradigm \cite{cai2025survey}, we adopt an alternative strategy: increasing the number of learnable base prompts. Similar to multiple experts in MoE, we construct a composite prompt by combining several base prompts, with their weights adaptively determined based on the characteristics of the adversarial image. This approach enables the generation of more diverse and expressive text features. Moreover, since each base prompt remains short and is easier to optimize, leading to improved adversarial robustness. A preliminary experiment has empirically verified our idea that increasing the number of prompts can indeed enhance robustness more effectively than simply extending the length of the prompts.



Based on the above consideration, in this paper, we propose an adversarial prompt tuning method named \textbf{Mixture of Adversarial Prompt Tuning (MoAPT)} to enhance the adversarial robustness of VLMs. Specifically, we fix the parameters of the text and image encoders but only optimize adversarial mixture prompts. These text prompts pass through the text encoder and generate diverse individual text features. In addition, to enhance the adaptability, we propose a conditional text weight router based on image features to predict the weights of adversarial mixture prompts and aggregate them into a sample-specific mixture text feature, so as to adaptively align with the diverse adversarial image features. A series of experiments show that our MoAPT can achieve better accuracy and robustness than state-of-the-art methods on multiple different datasets. Meanwhile, MoAPT also shows better generalization across different datasets. Our contribution can be summarized as follows:
\begin{itemize}
\item We find that for adversarial text prompt tuning, increasing the number of learnable text prompts can achieve a better robustness than only increasing the length of learnable text prompts within a certain range of parameters. 
\item We propose a novel method named Adversarial Mixture Prompt Tuning (MoAPT), which applies adversarial mixture prompts to generate diverse individual text features, where each text feature can play its unique roles for different adversarial examples, thereby alleviating the overfitting phenomenon. 
\item We apply a conditional text weight router based on image features to predict the weights of different text features and obtain a sample-specific mixture text feature that has pretty adaptability to align with different adversarial image features. Furthermore, we theoretically verify the effectiveness of the weight router.
\item We empirically verify the effectiveness of MoAPT. Extensive experiments demonstrate that our MoAPT can outperform state-of-the-art methods against adversarial examples in adversarial robustness and generalization across different datasets.
\end{itemize}



\section{Related Work}

\begin{figure*}[t]
    \centering
    \includegraphics[width=0.85\textwidth]{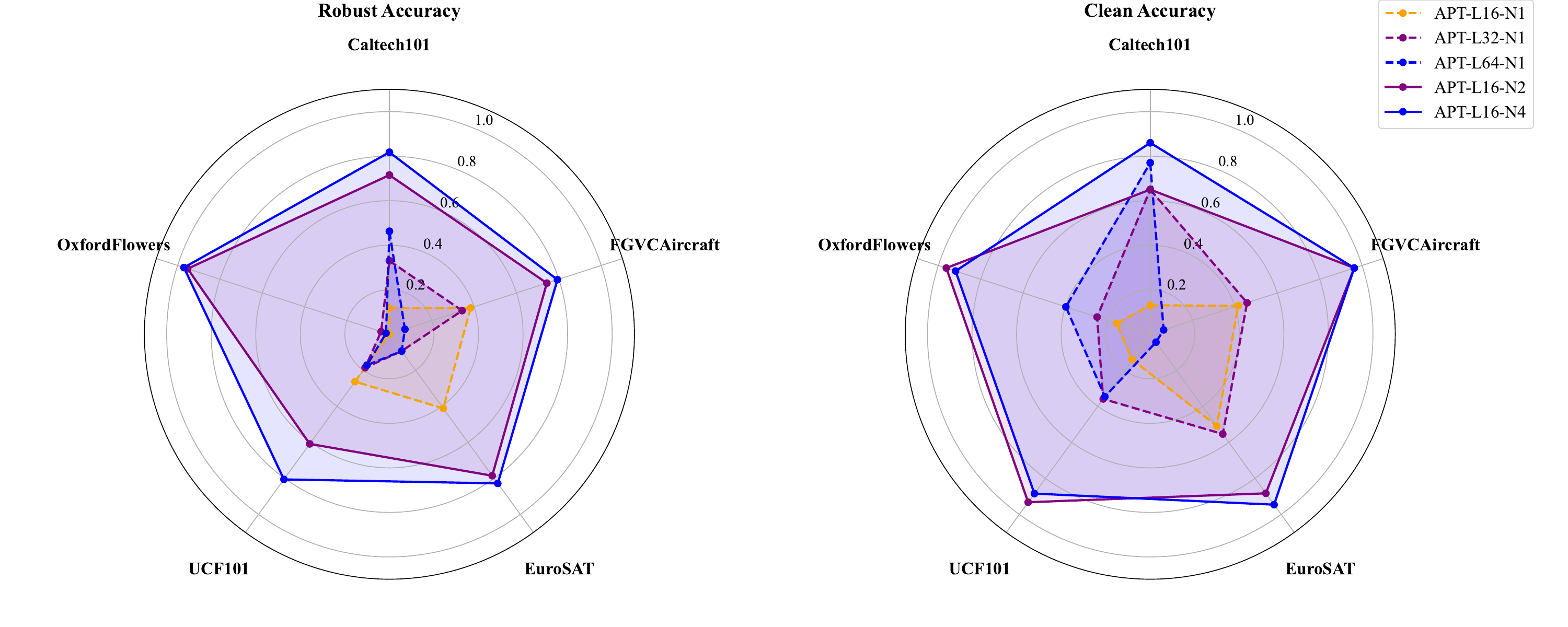}
    \caption{The performance of adversarial prompt tuning with different length and number on five datasets. ``APT-L{\textit{\textbf{m}}}-N{\textit{\textbf{k}}}'' denotes the APT with prompt length {\textit{\textbf{m}}} and prompt number {\textit{\textbf{k}}}. We find that increasing the number of prompts can enhance more robustness than increasing the prompt length (i.e., solid lines show better performance than dotted lines).}
    \label{fig:lidar}
\end{figure*}

\subsection{Prompt Tuning for Accuracy in VLMs}
Different from the methods of fine-tuning all model parameters, the prompt tuning method only fine-tunes the model's input prompts. Through a training process, a learnable prompt suitable for downstream tasks is obtained to replace the hand-crafted prompt, thereby improving the performance of the VLMs. The prompt tuning methods are originated from text model \cite{li2021prefix,liu2021p} and also have corresponding applications in visual models \cite{jia2022visual} and vision-language models \cite{khattak2023maple,zhou2022learning,zhou2022conditional}. CoOp \cite{zhou2022learning} first utilizes a learnable vector to replace the hand-crafted in Vision-Language Models. Based on CoOp, CoCoOp \cite{zhou2022conditional} is proposed by introducing a conditional Meta-net based on an image feature to generate an instance-adaptive vector and add it to the learnable vector. Some research also tries to apply multiple prompts in VLMs \cite{lu2022prompt,chen2022plot}.  Different from the above works, in this paper, we mainly focus on improving the adversarial robustness via optimizing multiple prompts and embedding it into the adversarial training framework, which has obvious differences.

\subsection{Adversarial Prompt Tuning in VLMs} 

Due to its excellent performance and low training cost, prompt tuning has been applied to improve the robustness of VLMs. \cite{chen2023visual} applies visual prompting to enhance the adversarial robustness. Furthermore, TeCoA \cite{mao2022understanding} and PMG-AFT \cite{wang2024pre} employ visual prompt tuning to improve the adversarial robustness of VLMs. AdvPT \cite{zhang2024adversarial} and APT \cite{li2024one} are proposed to apply text prompt tuning to further enhance the VLMs against image attacks. FAP \cite{zhou2024few} tries to enhance the robustness via bimodal tuning, while APD \cite{luo2024adversarial} further extends FAP into the adversarial distillation setting. To solve the insufficient generalization, \cite{wang2025tapt} applies Test-Time Adversarial Prompt Tuning (TAPT) to learns defensive bimodal (textual and visual) prompts during testing process. Different from the above research, this paper improves adversarial robustness through adversarial mixture prompt tuning during training process, which tries to solve the existing issue from another view but not conflict with each other.

\section{The Necessity of Mixture Prompts}

\subsection{Formulation of Adversarial Prompt Tuning}
 CoOp \cite{zhou2022learning} first applies the text prompt tuning in CLIP to improve the performance of downstream tasks, and \cite{zhang2024adversarial,li2024one} apply the adversarial prompt tuning in improving adversarial robustness, and the optimization goal of adversarial prompt tuning can be defined as follows:
\begin{align}
\label{sec3:eq-0}
    \mathop{\arg\min}_{t}\mathbb{E}_{(x,t,y)\sim \mathcal{D}}(\mathcal{L}(\Tilde{x},t,y;F_{\theta_{v}},F_{\theta_{t}})), 
\end{align}
 where $x$ and $t$ are the image and text pairs belong to the dataset $\mathcal{D}$. For the image classification task with $N$ classes, texts $t$ also contain $N$ different prompts: $\{t_1,t_2,\cdots,t_N\}$. $\Tilde{x}$ denotes adversarial examples. $y$ denotes the ground truth. $y_{in}$ indicates whether the image $x_i$ and text $t_{n}$ pair match, if the image $x_i$ and text $t_n$ match, $y_{in}$ is equal to 1, otherwise $y_{in}$ is equal to 0; the $F_{\theta_{v}}$ and $F_{\theta_{t}}$ are the image encoder and text encoder of CLIP. 
 
Meanwhile, as for the text $t$, a fixed text template, e.g., "a photo of a [CLASS]", is often directly used as the text input, and the maximum similarity between it and the input image is calculated to determine which class the image belongs to. \cite{zhang2024adversarial,li2024one} apply a learnable text prompt, which consists of the class context and a learnable context as follows:
\begin{align}
\label{sec3:eq-4}
t_{n} = [context_{front}][CLASS_{n}][context_{end}].
\end{align}
 
The image feature $z_{v}^{i}$ is generated by image encoder $F_{\theta_{v}}$ of input $\Tilde{x}_{i}$, the text feature $z_{t}^{n}$ is generated by text encoder $F_{\theta_{t}}$ of input $t_{n}$, which can be defined as follows:
\begin{align}
\label{sec3:eq-2}
\Tilde{z}_{v}^{i} = F_{\theta_{v}}(\Tilde{x}_{i}), z_{t}^{n} = F_{\theta_{t}}(t_{n}).
\end{align}

For the image classification task, Cross-Entropy loss is applied as the optimization function in APT \cite{li2024one}, which can be defined as follows:
\begin{align}
\label{sec3:eq-1}
    \mathcal{L}(\Tilde{x}_{i},t,y_{i};F_{\theta_{v}},F_{\theta_{t}}) = -\sum_{n=1}^Ny_{in}log\frac{exp(cos(\Tilde{z}_{v}^{i},z_{t}^{n}))}{\sum_{m=1}^N exp(cos(\Tilde{z}_{v}^{i},z_{t}^{m})}, 
\end{align}
where the cos similarity is applied to measure the alignment degree of the features, and applying the softmax operation can obtain the probability that the $\Tilde{x}_{i}$ aligns with the $z_{t}$.


\subsection{A Longer Prompt or More Prompts in APT?}

APT \cite{li2024one} extends the CoOp framework to enhance the robustness of VLMs against adversarial attacks. However, a single text prompt has potential generalization problems: when faced with complex adversarial examples, its parameters may struggle to adapt to the change. Therefore, we attempt to explore how to enhance the generalization of adversarial text prompt tuning from the perspectives of increasing length and increasing number.

To compare those two approaches, we keep the total prompt parameters the same. For example, we use a learnable prompt of length 64 to compare the robustness of 4 learnable prompts of length 16. We conducted experiments on five datasets and the results can be viewed in Figure \ref{fig:lidar}. And the experiments are based on APT \cite{li2024one}.

The results surprisingly reveal that, during adversarial prompt tuning, increasing the number of prompts is more effective than increasing their length. 
Specifically, we find that using 2/4 learnable adversarial prompts of length 16 achieves better adversarial robustness compared to using a single prompt of length 32/64, with an average improvement of 3.88\%/4.34\%. Notably, this setup also leads to an average performance gain of 4.56\%/6.43\% on clean samples. Furthermore, increasing the number of prompts further enhances adversarial robustness; When the number of prompts increases from 2 to 4, adversarial robustness improves by an additional 0.34\%. In contrast, merely increasing the prompt length yields no obvious robustness improvement.

We argue that after the total number of parameters reaches a certain level, continuing to increase the length of prompts will increase the difficulty of learning an ideal prompt. On the contrary, shorter prompts are relatively easier to learn, and adversarial mixture prompts can generate more diverse text features, which have more possibility to align with  adversarial examples. Therefore, increasing the number of prompts can further improve robustness compared with increasing prompt length.

\section{Mixture of Adversarial Prompt Tuning}

\label{sec:Methods}

\subsection{Overall Framework}

Based on the above findings, we argue that the generalization of adversarial robustness can be improved by adding adversarial mixture prompts. Therefore, we propose Mixture of Adversarial Prompt Tuning (MoAPT) to further improve the adversarial robustness of the VLMs. Here our optimization goal can be formulated as follows:
\begin{align}
\label{sec4:eq-1}
    \mathop{\arg\min}_{T_{m},\theta_{w}}\mathbb{E}_{(x,t,y)\sim \mathcal{D}}(\mathcal{L}_{MoAPT}(\Tilde{x},T_{m},y;F_{\theta_{v}},F_{\theta_{t}},F_{\theta_{w}})), 
\end{align}
where $\mathcal{L}_{MoAPT}$ denotes the optimization loss function of our MoAPT, and $T_{m}$ denotes the adversarial mixture prompts, $F_{\theta_{w}}$ denotes the conditional prompt weight router.
As for the adversarial examples $\Tilde{x}$, we follow the ``on-the-fly'' setting in \cite{li2024one}, where the attacker can access all the parameters of the VLMs including the adversarial mixture prompts but can only apply adversarial perturbations to the image $x$. And the adversarial examples $\Tilde{x}$ can be formulated as follows:
\begin{align}
\label{sec4:eq-2}
\Tilde{x}=\mathop{argmax}\limits_{||\Tilde{x}-x|| \leq \epsilon} \mathcal{L}_{MoAPT}(\Tilde{x},T_{m},y;F_{\theta_{v}},F_{\theta_{t}},F_{\theta_{w}})),
\end{align}
where $\epsilon$ denotes the maximum perturbation scale. It should be mentioned that the ``on-the-fly'' setting is closer to the adversarial examples in \cite{szegedy2013intriguing,madry2017towards}, which can access all parameters of the model and only modify the images. For the evaluation against adversarial attacks, we also follow this type of setting.

\subsection{Adversarial Mixture Prompts}

Assume adversarial mixture prompts $T_{m}$ have $K$ total of prompts, which can be defined as follows:
\begin{align}
\label{sec4:eq-0}
T_{m} = \{t^1,t^2,\cdots,t^K\},
\end{align}
where $t^k$ denotes the $k$-th learnable adversarial text prompt, which includes $N$ class text prompt: $\{t_1^k,t_2^k,\cdots,t_N^k\}$. Following \cite{zhou2022learning,li2024one}, the $CLASS_n$ context in each $t^k$ is represented by a sequence of class-specific vectors, and the learnable contexts are defined in the word embedding space, then $t^k$ for class $n$ can be formulated as follows:
\begin{align}
\label{sec4:eq-3}
t_n^k = [V ]_{1,n}^k ...[V ]_{M,n}^k[C_n], 
\end{align}
where the $M$ denotes the max length of learnable context. The position of $[C_n]$ can also be adjusted. Following \cite{li2024one}, we apply the end position as the default position. 


For adversarial mixture prompts, We first input different text prompts into the text encoder to obtain individual text features, then we aggregate these text features into a mixture text feature, which can be formulated as follows:
\begin{align}
\label{sec4:eq-4}
z_{t}^{n,i} = \sum_{k=1}^{K} \Tilde{w}_{k}^{i} F_{\theta_{t}}(t_{n}^{k}), 
\end{align}
where $\Tilde{w}_{k}^{i}$ denotes the weights of adversarial prompt $t^{k}$ for the adversarial examples $\Tilde{x}_i$, and $\Tilde{w}_{k}^{i}$ is irrelevant to the class (not effected by $n$). $z_{t}^{n,i}$ denotes the mixture text feature of $n$-th class for the adversarial examples $\Tilde{x}_i$. In this way, we can obtain adversarial mixture prompts with pretty diversity through adversarial training to defend against different adversarial examples.




\subsection{Conditional Prompt Weight Router}

Although adversarial mixture prompts can provide diverse adversarial text features, how to select those diverse features still needs to be solved when facing different adversarial examples.
Therefore, we focus to adjust the weights: $w_{k}^{i}$ in Eq. (\ref{sec4:eq-4}). The simplest approach is to convert $w_{k}^{i}$ to $1/K$. To cover diverse image adversarial examples, we propose the conditional prompt weight router, which can generate the image-specific multiple weights for the different adversarial prompts.

Here we design a light-weight network containing two full connection layers as the conditional prompt weight router to predict prompt weights $\Tilde{w}^{i}=\{\Tilde{w}_{1}^{i},\cdots,\Tilde{w}_{K}^{i}\}$ of image $x_i$. 
Initially, we obtain the adversarial image feature from the image encoder of VLMs $\Tilde{z}_{v}^{i}$,
then we apply the conditional prompt weight router to predict the different weights, which can be formulated as follows:
\begin{align}
\label{sec4:eq-6}
\Tilde{w}^{i} = softmax(F_{\theta_{w}}(\Tilde{z}_{v}^i)/\tau_{w}),
\end{align}
where $F_{\theta_{w}}$ denotes the conditional prompt weight router, the $z_v^i$ denotes the feature generated by image encoder of image $x_i$. The softmax operation can keep the sum of the weights as 1. The $\tau_{w}$ is applied to control the adjustment strength of the generated weight, while a smaller $\tau_{w}$ denotes stronger adjustment strength, and a larger $\tau_{w}$ denotes weaker adjustment strength, when $\tau_{w}$ approaches infinity, it will degenerate into $1/K$. With the assistance of an adaptive weight router, we can finally obtain a more generalizable and representative mixture text feature based on the image features, to further improve the adaptability to defend against different adversarial examples. Meanwhile, we also provide the Theorem \ref{corollary1} about our conditional prompt weight router.

\begin{algorithm}[t]  
  \caption{Training Process of MoAPT}  
  \label{algorithm:1}
  \begin{algorithmic}[1]   
   \Require {The train dataset $\mathcal{D}$, clean examples $x$ and adversarial examples $\Tilde{x}$, ground truth $y$, text encoder $F_{\theta_{t}}$ and image encoder $F_{\theta_{v}}$, adversarial mixture prompts with random initialization $T_{m} = \{t^1,t^2, \cdots,t^K\}$,  total class number $N$, condition prompt weight router $F_{\theta_{w}}$ with parameter $\theta_{w}$, the max training epochs $max$-$epoch$, the router temperature $\tau_{w}$}
     \For{$0$ to $max$-$epoch$} 
        \For{$Every~minibatch(x,t,y)~in~\mathcal{D}$}
             \State {  \small$\Tilde{x}=\mathop{argmax}\limits_{||\Tilde{x}-x|| \leq \epsilon} \mathcal{L}_{MoAPT}(\Tilde{x},T_{m},y;F_{\theta_{v}},F_{\theta_{t}},F_{\theta_{w}})$.}   
             \State { \small$\{z_{t,1},\cdots,z_{t,K}\}=  \{F_{\theta_{t}}(t^{1}),\cdots, F_{\theta_{t}}(t^{K})\}$.} 
            \For{$each ~x_i~ in~x$}            
          \State {\small  $\Tilde{z}_{v}^{i} =  F_{\theta_{v}}(\Tilde{x})$.}  
          \State { \small $\Tilde{w}^{i} = softmax(F_{\theta_{w}}(\Tilde{z}_{v}^{i})/\tau_{w})$.}   
          \State { \small $z_{t}^{n,i}= \sum_{k}^{K} \Tilde{w}_{k}^i z_{t,k}^{n}$.}  
          \EndFor  
        \State { \small $\theta_{w} = \theta_{w} - \eta \cdot \nabla_{\theta_{w}} \mathcal{L}_{MoAPT}$.}  
          \State { \small $T_{m} = T_{m} - \eta \cdot \nabla_{T_{m}} \mathcal{L}_{MoAPT}$.}  
          \EndFor  
    \EndFor  
  \end{algorithmic}  
\end{algorithm}

\newtheorem{thm3}{Theorem}
\begin{thm3}\label{corollary1}
Assume there are multiple different adversarial text prompts $T_{m} = \{t^1,t^2,\cdots,t^K\}$, and the corresponding error risk of $k$-th text prompt $t^k$ for  adversarial examples $\Tilde{x}$ is $\mathcal{R}(\Tilde{x},t^k,y)$, and the normalized prompt weights $\Tilde{w} = \{\Tilde{w}_{1},\Tilde{w}_{2},\cdots,\Tilde{w}_{K}\}$ are optimized to minimize the error risk expectation of adversarial example $\Tilde{x}$, we can obtain:
\begin{align}
\label{Theorem:eq0}
\mathbb{E}(\sum_{k}^K\Tilde{w}_{k}\mathcal{R}(\Tilde{x},t^k,y))\leq\mathbb{E}(\frac{1}{K}\sum_{k}^K\mathcal{R}(\Tilde{x},t^k,y)),
\end{align}
when there exists at least one pair $(i,j)$ exists $i\neq j$,  such that $\mathcal{R}(\Tilde{x},t^i,y))<\mathcal{R}(\Tilde{x},t^j,y))$, the strict inequality holds.
\end{thm3}

The proof of Theorem \ref{corollary1} can be viewed in Appendix \textcolor{red}{1.1}. Theorem \ref{corollary1} shows that conditional prompt weights can bring the smaller error expectation of the adversarial examples compared with the average error expectation of the adversarial examples, which further demonstrates the necessity and effectiveness of our conditional prompt weight router.

Then the entire process of our MoAPT can be viewed in Figure \ref{framework}, and the optimization loss function $\mathcal{L}_{MoAPT}$ can be defined as follows:
\begin{align}
\label{sec4:eq-7}
\mathcal{L}_{MoAPT} = -\sum_{n}^Ny_{in}log\frac{exp(cos(\Tilde{z}_{v}^{i},z_{t}^{n,i}))}{\sum_{n}^N exp(cos(\Tilde{z}_{v}^{i},z_{t}^{n,i}))},
\end{align}
and the final training process can be viewed in Algorithm \ref{algorithm:1}. It should be mentioned that to minimize the computational cost, we further decouple Eq. (\ref{sec4:eq-4}) and compute each text feature in advance for a minibatch. For each image, the final mixture text feature is obtained based on pre-computed text features without redundant calculation.

\section{Experiments}
\label{sec:Experiments}

\begin{table*}[t]
\centering
\caption{Robustness performance(\%) with all data training setting on 11 different datasets under maximum perturbation 4/255.}
\label{table:result16_and_all}
\renewcommand{\arraystretch}{1.4} 
\begin{adjustbox}{max width=\textwidth}
\begin{tabular}{c|cccccccccccc|c}
\hline
Methods & Metric &                    \multicolumn{1}{c}{\textbf{ImageNet}} & \multicolumn{1}{c}{\textbf{Caltech101}} & \multicolumn{1}{c}{\textbf{OxfordPets}} & \multicolumn{1}{c}{\textbf{Flowers102}} & \multicolumn{1}{c}{\textbf{Cars}} & \multicolumn{1}{c}{\textbf{FGVC}}     & \multicolumn{1}{c}{\textbf{DTD}} & \multicolumn{1}{c}{\textbf{SUN397}} & \multicolumn{1}{c}{\textbf{Food101}} & \multicolumn{1}{c}{\textbf{EuroSAT}} & \multicolumn{1}{c}{\textbf{UCF101}} & \multicolumn{1}{c}{\textbf{Average}} \\ \hline
                     & Clean & 39.84         & 77.44           & 61.49           & 30.37           & 10.33     & 7.02          & 27.13    & 31.98       & 21.70        & 20.31        & 36.16       & 33.07        \\
                       & PGD   & 10.27         & 44.02           & 14.28           & 8.73            & 0.92      & 0.48          & 11.17    & 5.86        & 3.19         & 9.25         & 6.24        & 10.40        \\
\multirow{-3}{*}{HEP}  & AA    & 7.24          & 39.92           & 11.01           & 6.41            & 0.62      & 0.06          & 9.52& 3.94        & 1.76         & 8.21         & 4.84        & 9.50         \\ \hline

&Clean & 48.84 & 84.63 & 62.25 & 67.19 & 1.07  & 0.99  & 22.05 & 48.91 & 39.89 & 78.89 & 12.11 & 42.44 \\
&PGD   & 5.78 & 51.36 & 15.51 & 34.51 & 0.91  & 0.99  & 9.99 & \textbf{17.48}  & \textbf{14.15}  & \textbf{51.70} & 4.57 & 18.81 \\
\multirow{-3}{*}{VPT \cite{mao2022understanding}}&AA    &  1.44 & 11.52 & 0.05  & 1.79  & 0.65  & 0.00  & 1.77  & 0.51  & 0.49  & 5.26 & 0.32 & 2.16\\ \hline
&Clean & 52.17 & 91.03 & 80.07 & 86.43 & 50.21 & 23.88 & 60.81 & 58.35 & 64.38 & 89.71 & 68.25 & 65.94 \\
&PGD   & 7.39 & 54.27 & 12.78 & 27.81 & 2.11  & 1.32  & 20.74 & 6.74  & 6.67  & 22.66 & 14.53 & 16.09 \\
\multirow{-3}{*}{FAP \cite{zhou2024few}}&AA    & 0.89 & 11.88 & 1.17  & 2.67  & 0.27  & 0.39  & 8.09  & 0.74  & 0.94  & 19.23 & 1.71 & 4.36  \\
\hline
                & Clean & 44.60         & 88.88           & 75.58           & 81.49           & 41.47     & 21.66         & 53.78    & 50.34       & 45.42        & 79.58        & 63.44       & 58.75        \\
                 & PGD   & 9.05          & 56.95           & 12.97           & 28.46           & 2.82      & 2.04          & 20.27    & 6.80        & 5.79         & 10.37        & 12.79       & 15.30        \\
\multirow{-3}{*}{AdvPT \cite{zhang2024adversarial}} & AA    & 7.02          & 55.13           & 11.07           & 24.73           & 1.62      & 1.26          & 18.79    & 5.50        & 4.06         & 9.22         & 10.73       & 13.56        \\ \hline
           & Clean & 41.48         & 88.32           & 72.58           & 80.88           & 37.42     & 20.49         & 52.19    & 47.29       & 35.32        & 68.67        & 59.00       & 54.88        \\
                 & PGD   & 12.57         & 63.65           & 24.56           & 44.90           & 8.93      & 7.05          & 26.24    & 13.15       & 13.11        & 24.51        & 21.89       & 23.69        \\

\multirow{-3}{*}{APT \cite{li2024one}}& AA    & 8.16& 61.01           & 16.43  & 38.61           & 3.92      & \textbf{3.33} & 22.40    & 8.06        & 7.32         & 29.79        & 16.39       & 19.58        \\ \hline
 
 & Clean                         & 42.30                                 & 87.38                    & 72.72& 82.34& 45.17                             & 20.58                                 & 53.43                            & 51.48                               & 38.98                                & 68.19                                & 60.27                               & 56.62                                \\
 
 & PGD                           & \textbf{12.62}                        & \textbf{65.03}                          & \textbf{24.78}                          & \textbf{46.81}                          & \textbf{12.16}                    & \textbf{7.56}                         & \textbf{28.49}                   & 13.92                      & 13.34                       & 37.70                       & \textbf{21.99}                      & \textbf{25.85}                       \\
 
\multirow{-3}{*}{\textbf{MoAPT (ours)}} & AA                            & \textbf{8.18}                         & \textbf{62.39}                          & \textbf{16.57}                          & \textbf{41.21}                          & \textbf{6.01}                     & 3.21& \textbf{25.41}                   & \textbf{9.91}                       & \textbf{8.03}                        & \textbf{34.56}                       & \textbf{17.29}                      & \textbf{21.16}                       \\ \hline
\end{tabular}
\end{adjustbox}
\end{table*}

\subsection{Experimental Setting}
\label{sec:5.1}
\textit{\textbf{Datasets.}} Following \cite{li2024one}, we  conduct our experiments on 11 high-resolution vision datasets: ImageNet \cite{deng2009imagenet}, Caltech101 \cite{fei2004learning}, OxfordPets \cite{parkhi2012cats}, StanfordCars \cite{krause20133d}, Flowers102 \cite{nilsback2008automated}, Food101 \cite{bossard2014food}, FGVCAircraft \cite{maji2013fine}, SUN397 \cite{xiao2010sun}, DTD \cite{cimpoi2014describing}, EuroSAT \cite{helber2019eurosat}, and UCF101 \cite{soomro2012ucf101}.  The 11 datasets were selected to establish a comprehensive benchmark, covering a wide range of vision tasks including generic object classification, scene recognition, action classification, fine-grained recognition, texture recognition, and satellite imagery analysis. They were split into training and test sets following the protocol of ~\cite{zhou2022learning}. 


\noindent \textit{\textbf{Models.}} Following the setting in \cite{li2024one}, we apply ViT-B/32 as our default selected backbone of image encoder, and select the model trained by a strong AT method TeCoA \cite{mao2022understanding} as our default optimized weight. 

\noindent \textit{\textbf{Baselines.}} Because our MoAPT is a text prompt tuning method, we mainly compare our method with some similar state-of-the-art methods: Hand Engineered Prompts (following the setting in \cite{li2024one}, see Appendix \textcolor{red}{1.3} for details), VPT \cite{mao2022understanding}, AdvPT \cite{zhang2024adversarial}, APT \cite{li2024one}, FAP \cite{zhou2024few}, where VPT is a visual prompt tuning method, AdvPT and APT are text prompt tuning method, FAP is the bi-modal tuning method.  Here we apply the HEP following the setting in \cite{li2024one}. 
Meanwhile, we change the setting of AdvPT into the setting of APT \cite{li2024one} for the sake of fair comparison. To ensure fairness, we apply the same backbone to further enhance the robustness for all the baselines.

\noindent \textit{\textbf{Evaluation Metric.}} Following the setting in \cite{li2024one}, we select two adversarial attacks, PGD attack \cite{madry2017towards} and AutoAttack \cite{croce2020reliable}. If without additional claim, we set the maximum perturbation $\epsilon$ of adversarial attacks to 4/255. For the PGD attack, we apply 100 iterations with a step $\epsilon/4$ following \cite{li2024one}. Meanwhile, we employ an ensemble attack, AutoAttack (AA) \cite{croce2020reliable}, which consists of four different attack methods: Auto-PGD (APGD), the Difference of Logits Ratio (DLR) attack, FAB-Attack \cite{croce2020minimally}, and the black-box Square Attack \cite{andriushchenko2020square}. All the methods are evaluated on the entire test test if without additional instruction. For the evaluation of ImageNet against Autoattack, we select the 5000 test set to reduce the calculation overhead following \cite{li2024one}, while conducting the AutoAttack on the entire test set is too expensive.

\noindent \textit{\textbf{Training settings.}} For each data set, we perform 16-shot and ``all'' training, where 16-shot denotes the 16 examples per class randomly sampled from the full training set for model training.   As for the training setting of our MoAPT, we train all the models with epoch 50 except ImageNet. Due to the high calculation overhead, we train on ImageNet with epoch 20 for ``all'' shot dataset and apply 100-shot similar to \cite{li2024one}. 
In the maximization of MoAPT, we generate the adversarial examples using 3 steps with a step size of 2$\epsilon$/3. 
Meanwhile, we set the prompt length to 16 and the number of prompts of our MoAPT to 8 except Sun397, Stanfordcars, and ImageNet. Due to the limitation of computing resources, for Sun397, Stanfordcars, and ImageNet, we set to prompt number as 3. Meanwhile, we set the hyper-parameter $\tau$ as 0.7. The corresponding discussion can be viewed in the Ablation Study. Meanwhile, we conduct the experiments on RTX 4090 except ImageNet, while ImageNet is conducted on A100.


\subsection{Robustness Performance}

\begin{table*}[t]
\centering
\caption{Robustness performance(\%) with 16-shot training setting on 11 different datasets under maximum perturbation 4/255.}
\label{table:result16}
\renewcommand{\arraystretch}{1.4} 
\begin{adjustbox}{max width=\textwidth}
\begin{tabular}{c|cccccccccccc|c}
\hline
Methods & Metric &                    \multicolumn{1}{c}{\textbf{ImageNet}} & \multicolumn{1}{c}{\textbf{Caltech101}} & \multicolumn{1}{c}{\textbf{OxfordPets}} & \multicolumn{1}{c}{\textbf{Flowers102}} & \multicolumn{1}{c}{\textbf{Cars}} & \multicolumn{1}{c}{\textbf{FGVC}}     & \multicolumn{1}{c}{\textbf{DTD}} & \multicolumn{1}{c}{\textbf{SUN397}} & \multicolumn{1}{c}{\textbf{Food101}} & \multicolumn{1}{c}{\textbf{EuroSAT}} & \multicolumn{1}{c}{\textbf{UCF101}} & \multicolumn{1}{c}{\textbf{Average}} \\ \hline
                                                                     & Clean  & 39.84                       & 77.44                                 & 61.49                                 & 30.37                                 & 10.33                                & 7.02                        & 27.13                                 & 31.98                                & 21.70                                & 20.31                                 & 36.16                        & 33.07            \\
                                                                     & PGD    & 10.27                       & 44.02                                 & 14.28                                 & 8.73                                  & 0.92                                 & 0.48                        & 11.17                                 & 5.86                                 & 3.19                                 & 9.25                                  & 6.24                         & 10.40            \\
\multirow{-3}{*}{HEP}                                                & AA     & 7.24                        & 39.92                                 & 11.01                                 & 6.41                                  & 0.62                                 & 0.06                        & 9.52                                  & 3.94                                 & 1.76                                 & 8.21                                  & 4.84                         & 9.50             \\ \hline
                                                                     & Clean  & 34.84                       & 76.92                                 & 3.38                                  & 41.49                                 & 3.38                                 & 1.05                        & 11.64                                 & 44.02                                & 1.16                                 & 6.70                                  & 2.11                         & 20.61            \\
                                                                     & PGD    & 3.13                        & 28.28                                 & 0.25                                  & 13.85                                 & 0.25                                 & 0.93                        & 0.71                                  & \textbf{13.39}                       & 0.08                                 & 0.00                                  & 0.13                         & 5.55             \\
\multirow{-3}{*}{VPT \cite{mao2022understanding}}   & AA     & 0.71                        & 0.30                                  & 0.14                                  & 0.20                                  & 0.14                                 & 0.00                        & 0.24                                  & 0.30                                 & 0.09                                 & 0.10                                  & 0.19                         & 0.22             \\ \hline
                                                                     & Clean  & 50.34                       & 89.85                                 & 76.09                                 & 76.24                                 & 43.68                                & 19.44                       & 50.29                                 & 56.24                                & 55.39                                & 64.67                                 & 63.75                        & 58.73            \\
                                                                     & PGD    & 6.92                        & 49.77                                 & 11.28                                 & 17.45                                 & 1.67                                 & 1.32                        & 16.72                                 & 2.44                                 & 4.21                                 & 13.48                                 & 10.46                        & 12.34            \\
\multirow{-3}{*}{FAP \cite{zhou2024few}}            & AA     & 0.51                        & 8.92                                  & 1.74                                  & 1.46                                  & 0.18                                 & 0.21                        & 6.85                                  & 1.40                                 & 0.66                                 & 11.17                                 & 0.97                         & 3.10             \\ \hline
                                                                     & Clean  & 43.09                       & 87.58                                 & 73.29                                 & 74.46                                 & 37.07                                & 19.92                       & 46.45                                 & 47.28                                & 36.05                                & 61.40                                 & 56.01                        & 52.96            \\
                                                                     & PGD    & 8.72                        & 50.67                                 & 12.84                                 & 20.99                                 & 2.69                                 & 2.07                        & 16.13                                 & 6.45                                 & 4.31                                 & 9.07                                  & 10.52                        & 13.13            \\
\multirow{-3}{*}{AdvPT \cite{zhang2024adversarial}} & AA     & 6.72                        & 49.53                                 & 10.47                                 & 16.89                                 & 1.69                                 & 0.96                        & 14.54                                 & 5.17                                 & 2.99                                 & 7.36                                  & 9.09                         & 11.40            \\ \hline
                                                                     & Clean  & 41.12                       & 86.29                                 & 67.29                                 & 76.41                                 & 31.6                                 & 20.31                       & 45.86                                 & 44.92                                & 30.39                                & 64.33                                 & 53.16                        & 51.06            \\
                                                                     & PGD    & 12.27                       & 56.75                                 & 19.98                                 & 37.52                                 & 7.70                                 & 6.15                        & 21.51                                 & 10.94                                & 7.90                                 & 25.54                        & 16.55                        & 20.26            \\
\multirow{-3}{*}{APT \cite{li2024one}}              & AA     & \textbf{7.88}               & 53.43                                 & 13.46                                 & 32.20                                 & 3.37                                 & \textbf{2.64}               & 18.91                                 & 6.88                                 & 4.17                                 & 16.68                                 & 12.74                        & 15.67            \\ \hline
                                                                     & Clean  & 41.20                       & 87.14                                 & {\color[HTML]{000000} 69.58}          & {\color[HTML]{000000} 77.63}          & 38.54                                & 19.29                       & 47.75                                 & 47.32                                & 30.73                                & 57.06                                 & 54.09                        & 51.84            \\
                                                                     & PGD    & \textbf{12.38}              & \textbf{57.69}                        & {\color[HTML]{000000} \textbf{21.29}} & {\color[HTML]{000000} \textbf{39.14}} & \textbf{9.29}                        & \textbf{6.63}               & \textbf{22.99}                        & 11.08                                & \textbf{8.46}                        & \textbf{28.96}                        & \textbf{18.61}               & \textbf{21.50}   \\
\multirow{-3}{*}{\textbf{MoAPT (ours)}}                                        & AA     & {\color[HTML]{000000} 7.84} & {\color[HTML]{000000} \textbf{55.05}} & {\color[HTML]{000000} \textbf{14.39}} & {\color[HTML]{000000} \textbf{34.02}} & {\color[HTML]{000000} \textbf{5.07}} & {\color[HTML]{000000} 2.52} & {\color[HTML]{000000} \textbf{20.04}} & {\color[HTML]{000000} \textbf{7.75}} & {\color[HTML]{000000} \textbf{4.60}} & {\color[HTML]{000000} \textbf{19.35}} & {\color[HTML]{000000} \textbf{14.14}} & \textbf{16.80}   \\ \hline
\end{tabular}
\end{adjustbox}
\end{table*}

We conduct a benchmark evaluation of our MoAPT and baseline approaches. Table \ref{table:result16_and_all} and Table \ref{table:result16} present the performance of various prompt methods in 11 datasets in both full-data and 16-shot training settings. Based on the results, MoAPT improves robustness by an average of 8.99\% and 11.33\% (PGD/AA) in the all-shot setting, and by 9.17\% and 8.82\% (PGD/AA) in the 16-shot setting. Furthermore, MoAPT achieves an average accuracy improvement of 5.60\% and 8.55\% (under all /16 shots training settings). It demonstrates strong adversarial robustness across various attacks while maintaining competitive accuracy.

Specifically, MoAPT consistently outperforms AdvPT and the best baseline APT in robustness and data efficiency. Under full-data training, MoAPT improves robustness over the best baseline by 2.16\% and 1.58\% (PGD/AA). In the 16-shot setting, MoAPT surpasses the baseline by 1.24\% and 1.13\% (PGD/AA), demonstrating its enhancement of APT’s performance under various attacks.
In contrast to MoAPT, VPT and FAP perform poorly in evaluating AA attacks, likely due to their lack of the generalization ability to unseen attacks, as seen in MoAPT. 

\begin{table*}[t]
\centering
\caption{Out-of-Distribution Robustness (\%) between APT and MoAPT cross 9 different datasets based on Caltech101 adversarial prompts. }
\label{table:crossdata_detail}
\renewcommand{\arraystretch}{1.0} 
\small
\begin{adjustbox}{max width=\textwidth}
\begin{tabular}{c|clllllllll|l}
\hline
Method & Metric                                                               & \multicolumn{1}{c}{\textbf{OxfordPets}} & \multicolumn{1}{c}{\textbf{Flowers102}} & \multicolumn{1}{c}{\textbf{Cars}} & \multicolumn{1}{c}{\textbf{FGVC}}     & \multicolumn{1}{c}{\textbf{DTD}}       & \multicolumn{1}{c}{\textbf{SUN397}} & \multicolumn{1}{c}{\textbf{Food101}}   & \multicolumn{1}{c}{\textbf{EuroSAT}}   & \multicolumn{1}{c}{\textbf{UCF101}} & \multicolumn{1}{c}{\textbf{Average}} \\ \hline
                                                            & Clean & 29.95           & 14.01           & 8.10& 1.83          & \textbf{16.84} & 14.19       & \textbf{15.52} & \textbf{12.81} & 19.51       & 14.75\\
                                                            & PGD   & \textbf{10.06}  & 3.17            & 0.87      & 0.36          & \textbf{7.74}  & 2.25        & 1.43           & 1.14& 3.49        & 3.39\\
\multirow{-3}{*}{APT}& AA    & \textbf{8.67}   & 2.15            & 0.39      & 0.33& 6.74           & 1.63        & 0.77           & 0.58           & 2.72        & 2.66\\ \hline
 
                                    & Clean                         & \textbf{45.08}                          & \textbf{16.00}& \textbf{13.01}                    & \textbf{2.34}                         & 16.02                                  & \textbf{20.17}                      & 15.3                                   & 11.43                                  & \textbf{26.20}& \textbf{18.39}\\
 
                                    & PGD                           & 9.10& \textbf{4.02}                           & \textbf{1.02}                     & \textbf{0.81}                         & 7.68& \textbf{4.10}& \textbf{2.24}                          & \textbf{10.67}                         & \textbf{4.34}                       & \textbf{4.89}\\
 
\multirow{-3}{*}{\textbf{MoAPT(ours)}}& AA                            & 7.14& \textbf{2.84}                           & \textbf{0.53}                     & \textbf{0.75}                         & \textbf{6.97}                          & \textbf{4.02}                       & \textbf{1.31}                          & \textbf{10.57}                         & \textbf{3.09}                       & \textbf{4.14}\\ \hline
\end{tabular}
\end{adjustbox}
\end{table*}

Meanwhile, we test the out-of-distribution robustness across different datasets. We apply the APT and MoAPT trained on Caltech101 with all-data training setting as source models and evaluate the adversarial robustness in different datasets, including OxfordPets, OxfordFlowers, StanfordCars, FGVCAircraft, Sun397, DTD, Food101, EuroSAT, and UCF101, and the average results are reported in Table \ref{table:crossdata_detail}. From the result, we find that MoAPT has better robust generalization compared with APT. Specifically, MoAPT has a 1.5\% and 1.48\% robustness improvement against PGD and AA attacks compared with APT, showing that MoAPT has better performance in dealing with diverse adversarial examples even in unseen classes.  MoAPT also outperforms in different backbone models, see Appendix \textcolor{red}{1.4} for details.

\subsection{Trade-off between Accuracy and Robustness}
\begin{figure}[t]
    \centering
    \includegraphics[width=0.48\textwidth]{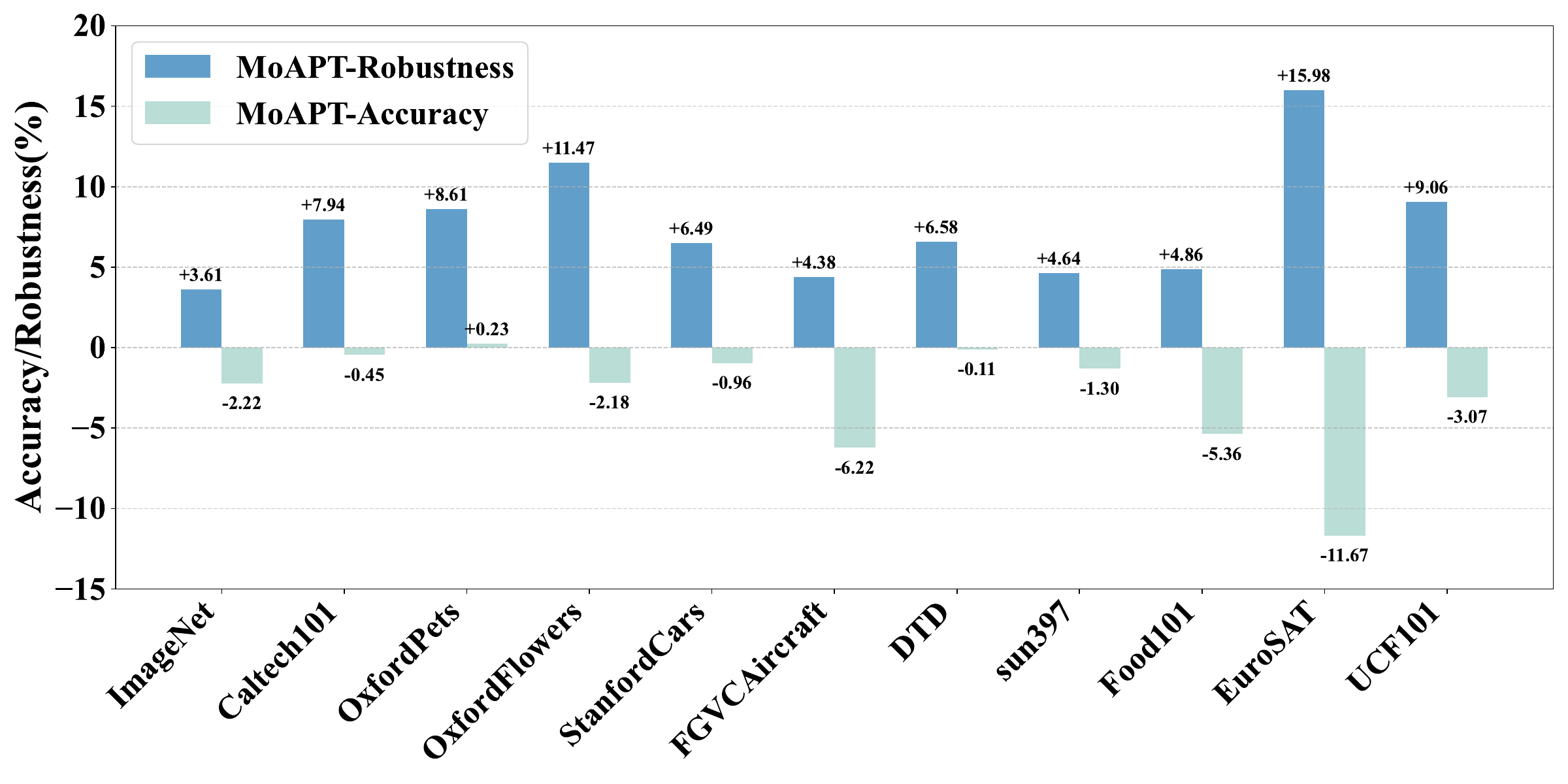}
    \caption{Trade-off between Accuracy and Robustness ($M=16$).}
    \label{trade-off}
\end{figure}

As shown in Fig.~\ref{trade-off}, we compare the performance improvement per dataset of our adversarially-trained prompt over the standard-trained prompt for unified context. Most adversarially trained vision models tend to improve robustness at the expense of accuracy, and adversarially trained prompts also exhibit this trade-off, which is expected. More importantly, we observe that for most datasets, the gain in robustness outweighs the drop in accuracy. Specifically, MoAPT improves adversarial robustness by an average of +7.60\%, while incurring only a modest drop of -3.03\% in accuracy. For instance, on OxfordPets, robustness increases significantly by +8.61\%, with a slight gain of +0.23\% in accuracy. These results suggest that our method achieves a relatively favorable trade-off between accuracy and robustness.

\subsection{Ablation Study}
To verify the effectiveness of MoAPT, we conduct a set of ablation studies. We conduct the experiment in the Caltech101 with the 16-shot setting. All the setting keep the same with the default setting if without additional instructions.

\begin{table}[ht]   
\centering
\caption{Ablation Study towards different components.}
\label{table:ablation1}
\scalebox{0.9}  {
\begin{tabular}{c|cccc}
\hline
Component      & Clean  & PGD & AA          \\  \hline 
Baseline  & 86.29 & 56.75 & 53.43      \\
Baseline+Mixture  &  87.06 & 57.36 & 54.60     \\
Baseline+Mixture+Router   & \textbf{87.14} & \textbf{57.69} & \textbf{55.05}  \\ \hline
\end{tabular}
}
\end{table}

\noindent \textit{\textbf{Effects of Different Components.}}
We conduct ablation studies on different components. Starting from the single adversarial cue fine-tuning baseline, we first add adversarial mixture prompts, and then further incorporate the conditional prompt weight router. Table \ref{table:ablation1} reports the results on Caltech101, with others given in Appendix \textcolor{red}{1.2}.

The results confirm the contribution of each module in MoAPT. Introducing mixture prompts without the weight router yields a modest robustness gain over the baseline, while integrating the conditional prompt-weight router provides a further improvement of 1.06\%/1.19\% under PGD/AA attacks and a 0.99\% increase in clean accuracy. These findings indicate that the feature diversity introduced by adversarial mixture prompts and the adaptive weighting enabled by the conditional router work in a complementary manner, jointly enhancing both the robustness and generalization of VLMs.


\begin{table}[ht] 
\centering
\caption{Ablation Study towards Prompt Number.}
\label{table:ablation2}
\scalebox{0.9}  {
\begin{tabular}{c|cccc}
\hline
Prompt number      & Clean  & PGD & AA          \\  \hline 
1  & 86.29 & 56.75 & 53.43   \\
2  &  86.13 & 56.98  & 53.66  \\
4  &  86.69 & 57.45 & 54.52   \\
6  &  87.22 & 57.04  & 54.32  \\
\textbf{8}   & 87.14 & \textbf{57.69} & \textbf{55.05} \\ 
10   & 87.34 & 56.80 & 54.40  \\ 
12   & 87.55 & 57.20 &  54.44 \\ 
\hline
\end{tabular}
}
\end{table}


\noindent \textit{\textbf{Selection of Prompt number.}}
We explore the selection of prompt numbers. 
We select the following text prompt number of our MoAPT  as 1, 2, 4, 6, 8, 10, 12, and the result can be viewed in Table \ref{table:ablation2}. 
From the results, when the number of prompts increases at the beginning (from number 1 to 8), the adversarial robustness of MoAPT will obviously increase. However, when it further increases (from number 8 to 12), the robustness remains basically unchanged. It can be explained that as the number of prompts increases, the difficulty of prompt optimization also increases. Thus, we select the prompt number 8 as the default setting. 
\begin{table}[ht] 
\centering
\caption{Ablation Study towards Hyper-parameter $\tau$.}
\label{table:ablation3}
\scalebox{0.9}  {
\begin{tabular}{c|cccc}
\hline
$\tau$      & Clean  & PGD  & AA         \\  \hline 
0.3  & 85.72 & 55.98 & 53.18     \\
0.5  & 86.86 & 57.93 & 54.32      \\
\textbf{0.7}  & 87.14 & 57.69 & \textbf{55.05}     \\
0.9   & 87.14 & 57.32 & 54.69  \\ 
1.1   & 87.05 & \textbf{57.77} & 54.56   \\ 
\hline
\end{tabular}
} 
\end{table}

\noindent \textit{\textbf{Selection of Hyper-parameter $\tau$.}}
The temperatures $\tau$ can control the adjustment strength of the conditional prompt weight router. 
While smaller $\tau$ means larger adjustment strength, larger $\tau$ means smaller adjustment strength.
We select the following $\tau$ of MoAPT as 0.3, 0.5, 0.7, 0.9, and 1.1, and the results can be found in Table \ref{table:ablation3}. Based on the experimental results, we select the Hyper-parameter $\tau$ to 0.7.

\subsection{Computational Cost} \label{limitation}
Despite adding multiple prompts, MoAPT still remains a parameter-efficient and highly competitive method as shown in Table \ref{table:Calculation overhead}. The inference memory and time costs of MoAPT are slightly higher than those of APT but are still lower than those of FAP, indicating that it maintains high inference efficiency while ensuring robustness.

\begin{table}[ht] 
\centering
\caption{Calculation Overhead. The results are conducted based on RTX 4090 in 16-shot setting of each epoch with Caltech101.}
\label{table:Calculation overhead}
\scalebox{0.85}  {
\begin{tabular}{c|ccccc}
\hline
Method      & VPT & APT & FAP & MoAPT        \\  \hline 
Training Memory Cost  & 6730M  & 2798M & 4204M  &  14384M \\
Training Time Cost  &  30s & 14s & 165s  & 60s  \\
\hline
Testing Memory Cost  & 2246M  & 5626M & 7478M  &  5838M \\
Testing Time Cost  &  5.00s & 8.42s & 14.01s  & 10.66s  \\
\hline
\end{tabular}
}
\end{table}

\section{Conclusion} \label{sec:Conclusion}

In this work, we focused on the overfitting problem of adversarial prompt tuning, and found that simply increasing the length of the text prompt led to the learning difficulty while increasing the number of prompts was more likely to improve the adversarial robustness of the VLMs. Based on the observation, we propose Mixture of Adversarial Prompt Tuning (MoAPT), which introduces adversarial mixture prompts to obtain more general text features, and proposes a conditional prompt weight router to further improve the adaptability of adversarial mixture prompts. Our theoretical analysis validates the effectiveness of the router. Extensive experiments demonstrate that MoAPT consistently improves in-distribution robustness and exhibits strong transfer robustness across diverse datasets.


{
    \small
    \bibliographystyle{ieeenat_fullname}
    \bibliography{main}
}


\end{document}


\maketitle

\section{Appendix} \label{sec:Appendix}
\subsection{The proof of Theorem 1}
\label{proof1}
Here we assume each text prompt keeps unchanged and only optimize the prompt weights $\Tilde{w} = \{\Tilde{w}_{1},\Tilde{w}_{2},\cdots,\Tilde{w}_{K}\}$  from initial weights $\Tilde{w}^0 = \{1/K,1/K,\cdots,1/K\}$. Then as for any adversarial example $\Tilde{x}^{'}$, without loss generalization, we can assmue a corresponding text prompt $t^k$ exist error risk as follows:
\begin{align}
\label{Theorem:eq1}
0\leq\mathcal{R}(\Tilde{x}^{'},t^k,y)\leq\mathcal{R}(\Tilde{x}^{'},t^1,y)\leq\cdots\leq\mathcal{R}(\Tilde{x}^{'},t^{k-1},y)  \notag \\ 
\leq\mathcal{R}(\Tilde{x}^{'},t^{k+1},y)\leq\cdots\leq\mathcal{R}(\Tilde{x}^{'},t^{K},y).
\end{align}

Thus based on the Gradient Descent operation, the expectation  of  corresponding weights can be formulated as follows:
\begin{align}
\label{Theorem:eq2}
\mathbb{E}(\Tilde{w}_{k}^{'}) = \frac{1}{K} - \eta \cdot \frac{\partial \frac{1}{K}\sum_{k}^K\mathcal{R}(\Tilde{x}^{'},t^k,y)}{\partial \Tilde{w}_{k}}, 
\end{align}
and here we assume that different weights are independent of each other, so we can obtain the $\mathbb{E}(\Tilde{w}_{K}^{'})$ as follows:
\begin{align}
\label{Theorem:eq3}
\mathbb{E}(\Tilde{w}_{k}^{'}) = \frac{1}{K} - \eta \cdot \mathcal{R}(\Tilde{x}^{'},t^k,y). 
\end{align}
Then, we apply softmax to each $\mathbb{E}(\tilde{w}_k^{'})$, and denote the result of applying softmax to $\mathbb{E}(\tilde{w}_k^{'})$ is $\mathbb{E}(\tilde{w}_{k})$, then we get:
\begin{align}
\label{Theorem:eq4}
\mathbb{E}(\tilde{w}_{k}^{'})=\frac{e^{ \frac{1}{K} - \eta \cdot \mathcal{R}(\Tilde{x}^{'},t^k,y)}}{\sum_{j}^Ke^{\frac{1}{K}-\eta\cdot\mathcal{R}(\Tilde{x}^{'},t^j,y)}}. 
\end{align}
For simplicity, we let $\mathcal{R}_k=\mathcal{R}(\Tilde{x}^{'},t^k,y)$, so Eq. (\ref{Theorem:eq4})It can be further simplified to:
\begin{align}
\label{Theorem:eq5}
\mathbb{E}(\tilde{w}_{k})=\frac{e^{  - \eta \cdot \mathcal{R}_k}}{\sum_{j}^Ke^{-\eta\cdot\mathcal{R}_j}}. 
\end{align}
Then the entire error expectation of adversarial examples can be formulated as follows:
\begin{align}
\label{Theorem:eq6}
\mathbb{E}(\sum_{k}^K\tilde{w}_{k}\mathcal{R}_k)=\mathbb{E}(\frac{1}{K}\sum_{k}^K\mathcal{R}_k) + \sum_{k}^K (\mathbb{E}(\tilde{w}_{k})-\frac{1}{K})*\mathcal{R}_k. 
\end{align}
After substituting $\mathbb{E}(\tilde{w}_{k})$ into $\sum_{k}^K (\mathbb{E}(\tilde{w}_{k})-\frac{1}{K})*\mathcal{R}_k$, we have:
\begin{align}
\label{Theorem:eq7}
\sum_{k}^K (\mathbb{E}(\tilde{w}_{k})-\frac{1}{K})*\mathcal{R}_k=\sum_{k}^K(\frac{e^{-\eta\cdot\mathcal{R}_k}}{\sum_{j}^Ke^{-\eta\cdot\mathcal{R}_j}}-\frac{1}{K})\cdot\mathcal{R}_k.
\end{align}
We assume the following relation holds:
\begin{align}
\label{Theorem:eq8}
\sum_{k}^K(\frac{e^{-\eta\cdot\mathcal{R}_k}}{\sum_{j}^Ke^{-\eta\cdot\mathcal{R}_j}}-\frac{1}{K})\cdot\mathcal{R}_k\leq0,
\end{align}
the above inequality is equivalent to the following inequality:
\begin{align}
\label{Theorem:eq8-1}
\sum_{k}^K(e^{-\eta\cdot\mathcal{R}_k}-\frac{\sum_{j}^Ke^{-\eta\cdot\mathcal{R}_j}}{K})\cdot\mathcal{R}_k\leq0.
\end{align}
When $K=2$, without loss of generality, we assume that $\mathcal{R}_1\leq\mathcal{R}_2$, then we have:
\begin{align}
\label{Theorem:eq7-1}
\sum_{k}^K(e^{-\eta\cdot\mathcal{R}_k}-\frac{\sum_{j}^Ke^{-\eta\cdot\mathcal{R}_j}}{K})\cdot\mathcal{R}_k  \notag \\  =\frac{e^{-\eta\cdot\mathcal{R}_1}-e^{-\eta\cdot\mathcal{R}_2}}{2}\cdot(\mathcal{R}_1-\mathcal{R}_2).
\end{align}
Since $\mathcal{R}_1\leq\mathcal{R}_2$, So we have $\mathcal{R}_1-\mathcal{R}_2\leq0$ and $\frac{e^{-\eta\cdot\mathcal{R}_1}-e^{-\eta\cdot\mathcal{R}_2}}{2}\geq0$, which means when $K=2$, the assumption holds. Next, we assume that the assumption holds when $K=n$,  and when $K=n+1$, without loss of generality, suppose $\mathcal{R}_{n+1}$ satisfies:
\begin{align}
\label{Theorem:eq8-2}
\mathcal{R}_{n+1}\leq\mathcal{R}_1\leq\dots\leq\mathcal{R}_{n},
\end{align}
then we have:
\begin{align}
\label{Theorem:eq20}
&\sum_{k}^{n+1}(e^{-\eta\cdot\mathcal{R}_k}-\frac{\sum_{j}^{n+1}e^{-\eta\cdot\mathcal{R}_j}}{K})\cdot\mathcal{R}_k \notag \\
&=\sum_{k}^{n+1}e^{-\eta\cdot\mathcal{R}_k}\cdot\mathcal{R}_k-\frac{\sum_{j}^{n+1}e^{-\eta\cdot\mathcal{R}_j}}{n+1}\cdot\sum_{k}^{n+1}\cdot\mathcal{R}_k.
\end{align}
By the induction hypothesis, we have:
\begin{align}
\label{Theorem:eq21}
\sum_{k}^{n+1}e^{-\eta\cdot\mathcal{R}_k}\cdot\mathcal{R}_k&=\sum_{k}^ne^{-\eta\cdot\mathcal{R}_k}\cdot\mathcal{R}_k+e^{-\eta\cdot\mathcal{R}_{n+1}}\cdot\mathcal{R}_{n+1} \notag \\ &\leq\frac{\sum_{j}^ne^{-\eta\cdot\mathcal{R}_j}}{n}\cdot\sum_{k}^n\mathcal{R}_k+e^{-\eta\cdot\mathcal{R}_{n+1}}\cdot\mathcal{R}_{n+1},
\end{align}
meanwhile, we have:
\begin{align}
\label{Theorem:eq22}
& \frac{\sum_{j}^{n+1}e^{-\eta\cdot\mathcal{R}_j}}{n+1}\cdot\sum_{k}^{n+1}\cdot\mathcal{R}_k \notag \\ 
& =\frac{\sum_{j}^ne^{-\eta\cdot\mathcal{R}_j}}{n+1}\cdot\sum_{k}^n\mathcal{R}_k+\frac{\sum_{j}^ne^{-\eta\cdot\mathcal{R}_j}}{n+1}\cdot\mathcal{R}_{n+1} \notag \\
&+\frac{e^{-\eta\cdot\mathcal{R}_{n+1}}}{n+1}\cdot\sum_{k}^n\mathcal{R}_k+\frac{e^{-\eta\cdot\mathcal{R}_{n+1}}}{n+1}\cdot\mathcal{R}_{n+1}.
\end{align}

Substituting Eq. (\ref{Theorem:eq22})  and Eq. (\ref{Theorem:eq21})  into Eq. (\ref{Theorem:eq20}), we can obtain:
\begin{align}
\label{Theorem:eq23}
&\sum_{k}^{n+1}(e^{-\eta\cdot\mathcal{R}_k}-\frac{\sum_{j}^{n+1}e^{-\eta\cdot\mathcal{R}_j}}{n+1})\cdot\mathcal{R}_k \notag  \\ 
&\leq[(\frac{\sum_{j}^ne^{-\eta\cdot\mathcal{R}_j}}{n(n+1)}\cdot\sum_{k}^n\mathcal{R}_k-\frac{e^{-\eta\cdot\mathcal{R}_{n+1}}}{n+1}\cdot\sum_{k}^n\mathcal{R}_k) \notag \\ 
&\notag+(\frac{n\cdot e^{-\eta\cdot\mathcal{R}_{n+1}}}{n+1}\cdot\mathcal{R}_{n+1}-\frac{\sum_{j}^ne^{-\eta\cdot\mathcal{R}_j}}{n+1}\cdot\mathcal{R}_{n+1})] \notag \\ 
&\notag=\frac{1}{n+1}\cdot(\frac{\sum_{j}^ne^{-\eta\cdot\mathcal{R}_j}}{n}-e^{-\eta\cdot\mathcal{R}_{n+1}})\cdot(\sum_{k}^n\mathcal{R}_k-n\cdot\mathcal{R}_{n+1}).
\end{align}

Since we have previously assumed that $\mathcal{R}_{n+1}\leq\mathcal{R}_1\leq\dots\leq\mathcal{R}_{n}$, so we can deduce that:
\begin{align}
\sum_{k}^n\mathcal{R}_k-n\cdot\mathcal{R}_{n+1}\geq0,
\end{align}

and:
\begin{align}
\label{Theorem:eq8-4}
\frac{\sum_{j}^ne^{-\eta\cdot\mathcal{R}_j}}{n}-e^{-\eta\cdot\mathcal{R}_{n+1}}\leq0.
\end{align}
Therefore, we can conclude that:
\begin{align}
\label{Theorem:eq30}
\sum_{k}^{n+1}(e^{-\eta\cdot\mathcal{R}_k}-\frac{\sum_{j}^{n+1}e^{-\eta\cdot\mathcal{R}_j}}{K})\cdot\mathcal{R}_k\leq0.
\end{align}
Therefore, by mathematical induction, we can conclude that the following holds for any K:
\begin{align}
\label{Theorem:eq32}
\sum_{k}^K(e^{-\eta\cdot\mathcal{R}_k}-\frac{\sum_{j}^Ke^{-\eta\cdot\mathcal{R}_j}}{K})\cdot\mathcal{R}_k\leq0.
\end{align}
Substituting Eq. (\ref{Theorem:eq32}) into Eq. (\ref{Theorem:eq6}), we obtain:
\begin{align}
\label{Theorem:eq33}
\mathbb{E}(\sum_{k}^K\tilde{w}_{k}\mathcal{R}_k)\leq\mathbb{E}(\frac{1}{K}\sum_{k}^K\mathcal{R}_k). 
\end{align}

Furthermore, form the above proof, we can deduce that if there exists at least one pair $(i,j)$, $i\neq j$, such that $\mathcal{R}_i<\mathcal{R}_j$, we have:

\begin{align}
\label{Theorem:eq34}
\mathbb{E}(\sum_{k}^K\tilde{w}_{k}\mathcal{R}_k)<\mathbb{E}(\frac{1}{K}\sum_{k}^K\mathcal{R}_k). 
\end{align}

Then the Theorem 1 is proved.

\subsection{Details on Ablation Study}
As shown in Table \ref{table:ablation_detail}, we report ablation results on 10 target datasets. The results indicate that incorporating both mixture prompts and the router consistently improves performance across almost all datasets, with average gains of 0.85/1.36/1.24 under clean accuracy, PGD, and AA attacks, respectively. In contrast, removing the router leads to lower performance than the full combination, suggesting that the two components are complementary: together they maintain high accuracy while enhancing adversarial robustness.

\label{appendix_ablation}

\begin{table*}[t]
\centering
\caption{Detailed ablation results cross 10 different datasets. }
\label{table:ablation_detail}
\renewcommand{\arraystretch}{1.4} 
\begin{adjustbox}{max width=\textwidth}
\begin{tabular}{
>{\columncolor[HTML]{FFFFFF}}c |
>{\columncolor[HTML]{FFFFFF}}c 
>{\columncolor[HTML]{FFFFFF}}l 
>{\columncolor[HTML]{FFFFFF}}l 
>{\columncolor[HTML]{FFFFFF}}l 
>{\columncolor[HTML]{FFFFFF}}l 
>{\columncolor[HTML]{FFFFFF}}l 
>{\columncolor[HTML]{FFFFFF}}l 
>{\columncolor[HTML]{FFFFFF}}l 
>{\columncolor[HTML]{FFFFFF}}l 
>{\columncolor[HTML]{FFFFFF}}l 
>{\columncolor[HTML]{FFFFFF}}l 
>{\columncolor[HTML]{FFFFFF}}l }
\hline
Method                                                & Metric & \multicolumn{1}{c}{\cellcolor[HTML]{FFFFFF}\textbf{Caltech101}} & \multicolumn{1}{c}{\cellcolor[HTML]{FFFFFF}\textbf{OxfordPets}} & \multicolumn{1}{c}{\cellcolor[HTML]{FFFFFF}\textbf{Flowers102}} & \multicolumn{1}{c}{\cellcolor[HTML]{FFFFFF}\textbf{Cars}} & \multicolumn{1}{c}{\cellcolor[HTML]{FFFFFF}\textbf{FGVC}} & \multicolumn{1}{c}{\cellcolor[HTML]{FFFFFF}\textbf{DTD}} & \multicolumn{1}{c}{\cellcolor[HTML]{FFFFFF}\textbf{SUN397}} & \multicolumn{1}{c}{\cellcolor[HTML]{FFFFFF}\textbf{Food101}} & \multicolumn{1}{c}{\cellcolor[HTML]{FFFFFF}\textbf{EuroSAT}} & \multicolumn{1}{c}{\cellcolor[HTML]{FFFFFF}\textbf{UCF101}} & \multicolumn{1}{c}{\cellcolor[HTML]{FFFFFF}\textbf{Average}} \\ \hline
\cellcolor[HTML]{FFFFFF}                                       & Clean  & 86.29                                                           & 67.29                                                           & 76.41                                                           & 31.6                                                      & \textbf{20.31}                                            & 45.86                                                    & 44.92                                                       & 30.39                                                        & \textbf{64.33}                                               & \multicolumn{1}{l|}{\cellcolor[HTML]{FFFFFF}53.16}          & 52.06                                                        \\
\cellcolor[HTML]{FFFFFF}                                       & PGD    & 56.75                                                           & 19.98                                                           & 37.52                                                           & 7.7                                                       & 6.15                                                      & 21.51                                                    & 10.94                                                       & 7.9                                                          & \textbf{25.54}                                               & \multicolumn{1}{l|}{\cellcolor[HTML]{FFFFFF}16.55}          & 21.05                                                        \\
\multirow{-3}{*}{\cellcolor[HTML]{FFFFFF}Baseline}         & AA     & 53.43                                                           & 13.46                                                           & 32.20                                                           & 3.37                                                      & \textbf{2.64}                                             & 18.91                                                    & 6.88                                                        & 4.17                                                         & 16.68                                                        & \multicolumn{1}{l|}{\cellcolor[HTML]{FFFFFF}12.74}          & 16.45                                                        \\ \hline
\cellcolor[HTML]{FFFFFF}                                       & Clean  & 86.45                                                           & \textbf{69.61}& 76.45                                                           & 34.56                                                     & 17.88                                                     & 46.39                                                    & 45.51                                                       & 30.2                                                         & 56.89                                                        & \multicolumn{1}{l|}{\cellcolor[HTML]{FFFFFF}\textbf{55.22}} & 51.92                                                        \\
\cellcolor[HTML]{FFFFFF}                                       & PGD    & 57.24                                                           & 20.66                                                           & 37.61                                                           & 7.8                                                       & 6.21                                                      & 21.93                                                    & 11.01                                                       & 8.14                                                         & 24.3                                                         & \multicolumn{1}{l|}{\cellcolor[HTML]{FFFFFF}18.55}          & 21.35                                                        \\
\multirow{-3}{*}{\cellcolor[HTML]{FFFFFF}Baseline+Mixture} & AA     & 53.75                                                           & 14.01                                                           & 32.23                                                           & 3.55                                                      & 2.49                                                      & 18.79                                                    & 6.89                                                        & 4.38                                                         & 15.53                                                        & \multicolumn{1}{l|}{\cellcolor[HTML]{FFFFFF}13.35}          & 16.5                                                         \\ \hline
\cellcolor[HTML]{FFFFFF}                                       & Clean  & \textbf{87.14}                                                  & 69.58& \textbf{77.63}& \textbf{38.54}                                            & 19.29                                                     & \textbf{47.75}                                           & \textbf{47.32}                                              & \textbf{30.73}                                               & 57.06                                                        & \multicolumn{1}{l|}{\cellcolor[HTML]{FFFFFF}54.09}          & \textbf{52.91}                                               \\
\cellcolor[HTML]{FFFFFF}                                       & PGD    & \textbf{57.69}                                                  & \textbf{21.29}& \textbf{39.14}& \textbf{9.29}                                             & \textbf{6.63}                                             & \textbf{22.99}                                           & \textbf{11.08}                                              & \textbf{8.46}                                                & \textbf{24.72}                                               & \multicolumn{1}{l|}{\cellcolor[HTML]{FFFFFF}\textbf{18.61}} & \textbf{22.41}\\
\multirow{-3}{*}{\cellcolor[HTML]{FFFFFF}Baseline+Mixture+Router}& AA     & \textbf{55.05}                                                  & \textbf{14.39}& \textbf{34.02}& \textbf{5.07}                                             & 2.52                                                      & \textbf{20.04}                                           & \textbf{7.75}                                               & \textbf{4.6}                                                 & 19.35                                                        & \multicolumn{1}{l|}{\cellcolor[HTML]{FFFFFF}\textbf{14.14}} & \textbf{17.69}\\ \hline
\end{tabular}
\end{adjustbox}
\end{table*}



\subsection{Details of the Dataset and Hand-Engineered Prompt}
\label{hep and datasets}
The 11 datasets were selected to establish a comprehensive benchmark, covering a wide range of vision tasks including generic object classification, scene recognition, action classification, fine-grained recognition, texture recognition, and satellite imagery analysis. They were split into training and test sets following the protocol of ~\cite{zhou2022learning}. Table \ref{table: dataset and hep} reports the number of categories in each dataset along with the corresponding inputs for the hand-engineered prompts.


\begin{table}[ht]
\centering
\caption{Dataset statistics and hand-crafted prompts}
\label{table: dataset and hep}
\resizebox{\columnwidth}{!}{%
\begin{tabular}{ccc}
\hline
Dataset      & Classes & Hand-Engineered Prompt                       \\ \hline
ImageNet     & 1000    & a photo of a {[}CLASS{]}                     \\
Caltech101   & 100     & a photo of a {[}CLASS{]}                     \\
OxfordPets   & 37      & a photo of a {[}CLASS{]}, a type of pet      \\
Flowers102   & 102     & a photo of a {[}CLASS{]}, a type of flower   \\
StanfordCars & 196     & a photo of a {[}CLASS{]}                     \\
FGVC         & 100     & a photo of a {[}CLASS{]}, a type of aircraft \\
DTD          & 47      & {[}CLASS{]} texture                          \\
SUN397       & 397     & a photo of a {[}CLASS{]}                     \\
Food101      & 101     & a photo of a {[}CLASS{]}, a type of food     \\
EuroSAT      & 10      & a centered satellite photo of a {[}CLASS{]}  \\
UCF101       & 101     & a photo of a person doing {[}CLASS{]}        \\ \hline
\end{tabular}%
}
\end{table}

\subsection{Generalization to Other Architectures}
As shown in Table \ref{table:resnet50}, our method uses the ResNet50 visual encoder. Our method outperforms the APT baseline method across various settings, achieving improvements of 1.6\%/2.17\% (PGD/AA). Furthermore, MoAPT achieves higher accuracy while improving robustness, demonstrating that our method maintains outstanding performance even when different backbone models are used.

\begin{table*}[t]
\centering
\caption{Detailed Performance for ResNet50 under maximum perturbation 1/255 and 4/255 on the Caltech101 Dataset. }
\label{table:resnet50}
\renewcommand{\arraystretch}{1.0} 
\small
\begin{adjustbox}{max width=\textwidth}
\begin{tabular}{c|ccccc}
\hline
Methods                         & Metric & \textbf{16shot(1/255)}                & \textbf{16shot(4/255)}                & \textbf{allshot(1/255)}               & \textbf{allshot(4/255)}               \\ \hline
                                         & Clean           & {\color[HTML]{000000} 86.86}          & {\color[HTML]{000000} 82.35}          & {\color[HTML]{000000} 89.29}          & {\color[HTML]{000000} 80.32}          \\
                                         & PGD             & {\color[HTML]{000000} 73.39}          & {\color[HTML]{000000} 32.25}          & {\color[HTML]{000000} 78.22}          & {\color[HTML]{000000} 44.02}          \\
\multirow{-3}{*}{APT}           & AA              & {\color[HTML]{000000} 72.37}          & {\color[HTML]{000000} 23.2}           & {\color[HTML]{000000} 77.12}          & {\color[HTML]{000000} 35.38}          \\ \hline
                                         & Clean           & {\color[HTML]{000000} 87.83}          & {\color[HTML]{000000} 83.45}          & {\color[HTML]{000000} 90.18}          & {\color[HTML]{000000} 83.85}          \\
                                         & PGD             & {\color[HTML]{000000} \textbf{74.89}} & {\color[HTML]{000000} \textbf{34.16}} & {\color[HTML]{000000} \textbf{79.43}} & {\color[HTML]{000000} \textbf{45.8}}  \\
\multirow{-3}{*}{\textbf{MoAPT   (ours)}} & AA              & {\color[HTML]{000000} \textbf{73.79}} & {\color[HTML]{000000} \textbf{26.21}} & {\color[HTML]{000000} \textbf{78.58}} & {\color[HTML]{000000} \textbf{38.17}} \\ \hline
\end{tabular}
\end{adjustbox}
\end{table*}

\subsection{Effect of the perturbation budget}
To examine whether the robustness behavior changes under different attack strengths, we evaluate PGD attacks with multiple perturbation budgets on Caltech101 while keeping the same prompt settings as in the main experiments. Table \ref{tab:epsilon_caltech101} reports the accuracy under $\epsilon$ ranging from 4/255 to 16/255. As expected, increasing the perturbation budget leads to a monotonic degradation in accuracy, from 57.69\% at $\epsilon=4/255$ to 1.54\% at $\epsilon=16/255$. Although the absolute robustness decreases with stronger perturbations, the overall robustness trend remains consistent across different $\epsilon$ values. This indicates that the behaviour observed in our main analysis is not tied to a specific perturbation budget and remains stable under different attack strengths.

\begin{table*}[ht]
\centering
\small
\caption{Robust Accuracy (\%) under different PGD perturbation budgets on Caltech101.}
\label{tab:epsilon_caltech101}
\begin{tabular}{c|cccccc}
\hline
$\epsilon$ & 4/255 & 8/255 & 10/255 & 12/255 & 14/255 & 16/255 \\ \hline
Robust Accuracy (\%)      & 57.69 & 21.20 & 9.94 & 4.75 & 2.56 & 1.54 \\ \hline
\end{tabular}
\end{table*}

{
    \small
    \bibliographystyle{ieeenat_fullname}
    \bibliography{main}
}
